\documentclass[runningheads]{llncs}
\usepackage{graphicx}
\usepackage{amssymb}
\usepackage{amsmath}
\usepackage{hyperref}

\begin{document}
\title{Automated Kantian Ethics: A Faithful Implementation}
\titlerunning{Automated Kantian Ethics}
%
\author{Lavanya Singh\orcidID{0000-0001-7791-3172
} }
\authorrunning{L. Singh}
%
\institute{Harvard University, Cambridge MA 02138, USA \\
\email{lavanyasingh2000@gmail.com}}
\maketitle              
\begin{abstract}
As we grant artificial intelligence increasing power and independence in contexts like healthcare, policing, and driving, AI faces moral dilemmas but lacks the tools to solve them. Warnings from regulators, philosophers, and computer scientists about the dangers of unethical artificial intelligence have spurred interest in automated ethics—i.e., the development of machines that can perform ethical reasoning. However, prior work in automated ethics rarely engages with philosophical literature. Philosophers have spent centuries debating moral dilemmas so automated ethics will be most nuanced, consistent, and reliable when it draws on philosophical literature. In this paper, I present an implementation of automated Kantian ethics that is faithful to the Kantian philosophical tradition. I formalize Kant's categorical imperative in Dyadic Deontic Logic, implement this formalization in the Isabelle/HOL theorem prover, and develop a testing framework to evaluate how well my implementation coheres with expected properties of Kantian ethic.  My system is an early step towards philosophically mature ethical AI agents and it can make nuanced judgements in complex ethical dilemmas because it is grounded in philosophical literature. Because I use an interactive theorem prover, my system's judgements are explainable.

\keywords{Automated ethics  \and Kant \and Isabelle \and AI ethics.}
\end{abstract}
\section{Introduction}\label{intro}

AI is making decisions in increasingly important contexts, such as medical diagnoses and criminal sentencing, and must perform ethical reasoning to navigate the world responsibly. This ethical reasoning will be most nuanced and trustworthy when it is informed by philosophy. Prior work in building computers that can reason about ethics, known as automated ethics, rarely capitalizes on philosophical progress and thus often cannot withstand philosophical scrutiny. This paper presents an implementation\footnote{Source code can be found at \href{https://github.com/lsingh123/automatedkantianethics}{https://github.com/lsingh123/automatedkantianethics}.} of philosophically faithful automated ethics.

Faithfully automating ethics is challenging. Representing ethics using constraint satisfaction~\cite{csp} or reinforcement learning~\cite{util1} fails to capture most ethical theories. For example, encoding ethics as a Markov Decision Process assumes that ethical reward can be aggregated, a controversial idea~\cite{consequentialismsep}. Even once ethics is automated, context given to the machine, such as the description of an ethical dilemma, plays a large role in determining judgements. 

I implement automated Kantian ethical reasoning that is faithful to philosophical literature. I formalize Kant's moral rule in Dyadic Deontic Logic (DDL), a logic that can express obligation and permissibility~\cite{CJDDL}. I implement my formalization in Isabelle, an interactive theorem prover that can automatically generate proofs in user-defined logics~\cite{isabelle}. Finally, I use Isabelle to automatically prove theorems (such as, ``murder is wrong'') in my new logic. Because my system automates reasoning in a logic that represents Kantian ethics, it automates Kantian ethical reasoning. It can classify actions as prohibited, permissible or obligatory with minimal factual background. I make the following contributions:
\begin{enumerate}
\item In Section \ref{formalization}, I formalize  a philosophically accepted version of Kant's moral rule in DDL. 
\item In Section \ref{implementation}, I implement my formalization in Isabelle. My system can judge appropriately-represented actions and show the facts used in the proof.
\item In subsections 1 and 2 of Section \ref{joking}, I use my system to produce nuanced answers to two well-known Kantian ethical dilemmas. Because my system draws on Kantian literature, it can perform sophisticated moral reasoning. 
\item In Section \ref{tests}, I present a testing framework to evaluate how faithful my system is to philosophical literature. Tests show that my implementation outperforms two other formalizations of Kantian ethics. 
\end{enumerate}

\section{The Need for Faithful, Explainable Automated Ethics}\label{problem}

AI operating in high-stakes environments like policing and healthcare must make moral decisions. For example, self-driving cars may face the following moral dilemma: an autonomous vehicle approaching an intersection fails to notice pedestrians until it is too late to brake. The car can continue on its course, running over and killing three pedestrians, or it can swerve to hit a tree, killing its single passenger. While this example is (hopefully) not typical of the operation of a self-driving car, every decision that such an AI agent makes, from avoiding congested freeways to carpooling, is morally tinged. 

Machine ethicists recognize this need and have made theoretical (\cite{moralmachineonline,davenport,moralmachine,gabriel}) and practical progress in automating ethics (\cite{logicprogramming,biology,delphi,winfield}). Prior work in machine ethics using deontology (\cite{deon2,deon1}), consequentialism (\cite{util1,util2,cloos}), and virtue ethics (\cite{berberich}) rarely engages with philosophical literature, and so misses philosophers' insights. The example of the self-driving car is an instance of the trolley problem~\cite{foot}, in which a bystander watching a runaway trolley can pull a lever to kill one and save three. Decades of philosophical debate have developed nuanced answers to the trolley problem. AI's moral dilemmas are not entirely new, so solutions should draw on philosophical progress. The more faithful that automated ethics is to philosophy, the more trustworthy and nuanced it will be.

A lack of engagement with philosophical literature also makes automated ethics less explainable, as seen in the example of Delphi, which uses deep learning to make moral judgements based on a training dataset of human decisions~\cite{delphi}. Early versions of Delphi gave unexpected results, such as declaring that the user should commit genocide if it makes everyone happy~\cite{verge}. Because no explicit ethical theory underpins Delphi's judgements, we cannot determine why Delphi thinks genocide is obligatory. Machine learning approaches like Delphi often cannot explain their decisions. This reduces human trust in a machine's controversial ethical judgements. The high stakes of automated ethics require explainability to build trust and catch mistakes. 

\section{Automated Kantian Ethics}\label{myidea}

I present a faithful implementation of Kantian ethics, a testing framework to evaluate how well my implementation coheres with philosophical literature, and examples of my system performing sophisticated moral reasoning. 

I formalize Kant's moral rule in Dyadic Deontic Logic~\cite{CJDDL}. Deontic logic is a modal logic that can express obligation, or binding moral requirements. Modal logics include the necessitation operator $\Box$, where $\Box p$ is true at world $w$ if $p$ is true at all worlds that neighbor $w$~\cite{cresswell}. Modal logics also contain operators of propositional logic like $\neg, \wedge, \vee, \rightarrow$. Deontic logics replace the $\Box$ operator with an obligation operator $O$. I use Carmo and Jones's Dyadic Deontic Logic (DDL)~\cite{CJDDL}, which uses the dyadic obligation operator $O\{A \vert B\}$ to represent the sentence ``A is obligated in the context B.'' 

Because this work is an early step towards faithful automated ethics, I use Kantian ethics, a theory that is amenable to formalization. I do not argue that Kantian ethics is the best theory, but that it is the most natural to automate.\footnote{The full argument is in Appendix \ref{whykant}.} I automate the Formula of Universal Law (FUL), a version of Kant's moral rule that states that moral principles can be acted on by all people without contradiction. For example, if everyone falsely promises to repay a loan, lenders will stop offering loans, so not everyone can act on this principle, so it is prohibited.

Prior work by Benzmüller et. al.~\cite{logikey,BFP} implements DDL in Isabelle. I add the Formula of Universal Law as an axiom to their library. The resulting Isabelle theory can automatically generate proofs in a new logic that has the categorical imperative as an axiom. Because interactive theorem provers are designed to be interpretable, my system is explainable. Isabelle can list the facts used in a proof and construct human-readable proofs. In Section \ref{joking}, I use my system to generate sophisticated solutions to two ethical dilemmas. Because my system is faithful to philosophical literature, it produces nuanced judgements. 

I also contribute a testing framework that evaluates how well my formalization coheres with philosophical literature. I formalize expected properties of Kantian ethics as sentences in my logic and run the tests by using Isabelle to automatically find proofs or countermodels for the test statements. My system outperforms two other attempts at formalizing Kantian ethics~\cite{kroy}.

Given an action represented as a sentence in my logic, my system proves that it is morally obligatory, permissible, or prohibited. My system serves as one step towards philosophically sophisticated automated ethics. 

\section{Details}

\subsection{Formalizing the Categorical Imperative in DDL}\label{formalization}

The Formula of Universal Law reads, ``act only according to that maxim by which you can at the same time will that it should become a universal law''~\cite{groundwork}. To formalize this, I represent willing, maxims, and the FUL in DDL.

\subsubsection{Willing a Maxim}\label{maxim}

Kantian ethics evaluates ``maxims,'' which are ``the subjective principles of willing,'' or the principles that the agent understands themselves as acting on~\cite{groundwork}. I adopt O'Neill's view that a maxim includes the act, the circumstances, and the agent's purpose of acting or goal~\cite{actingonprinciple}. 

\begin{definition}[Maxim]
    A circumstance, act, goal tuple (C, A, G), read as ``In circumstances C, do act A for goal G.''
\end{definition}

For example, one maxim is ``When strapped for cash, falsely promise to repay a loan to get some easy money.'' A maxim includes an act and the circumstances\footnote{The inclusion of circumstances in a maxim raises the ``tailoring objection''~\cite{whatisamaxim,kantsethicalthought}, under which maxims are arbitrarily specified to pass the FUL. For example, the maxim ``When my name is John Doe, I will lie to get some easy money,'' passes the FUL but should be prohibited. One solution is to argue that the circumstance ``when my name is John Doe'' is not morally relevant, but this requires defining morally relevant circumstances. The difficulty in determining relevant circumstances and formulating a maxim is a limitation of my system and requires that future work develop heuristics to classify circumstances as morally relevant. 
} under which it should be performed. It must also include a goal because human activity, guided by a rational will, pursues ends that the will deems valuable~\cite{groundwork}. 

I define ``willing a maxim'' as adopting it as a principle to live by.

\begin{definition}[Willing]
For maxim $M = (C, A, G)$ and actor $s$,

$$will \, M \, s \equiv \forall w \, (C \longrightarrow A \, (s)) \, w$$

\noindent At all worlds $w$, if the circumstances hold at that world, agent $s$ performs act $A$.

\end{definition}

\noindent If I will the example maxim above about falsely promising to repay a loan, then whenever I need cash, I will falsely promise to repay a loan. 

\subsubsection{Practical Contradiction Interpretation}\label{praccon}

My project uses Korsgaard's canonical practical contradiction interpretation of the FUL~\cite{KorsgaardFUL,ebelsduggan}. 

The logical contradiction interpretation prohibits maxims that are impossible when universalized. Under this view, falsely promising is wrong because, in the universalized world, the practice of promising would end, so falsely promising would be impossible. This view cannot handle natural acts, like that of a mother killing her crying children so that she can get some sleep~\cite{dietrichson,KorsgaardFUL}. Universalizing this maxim does not generate a contradiction, but it is clearly wrong. Because killing is a natural act, it can never be impossible so the logical contradiction view cannot prohibit it.

As an alternative to the logical contradiction view, Korsgaard endorses the practical contradiction view, which prohibits maxims that are self-defeating, or ineffective, when universalized. By willing a maxim, an agent commits themselves to the maxim's goal, so they cannot rationally will that this goal be undercut. This can prohibit natural acts like that of the sleep-deprived mother: in willing the end of sleeping, she is willing that she is alive. If all mothers kill all loud children, then she cannot be secure in the possession of her life, because her mother could have killed her as an infant. Willing this maxim thwarts the end that she sought to secure. 

\subsubsection{Formalizing the FUL}\label{formalizingful}

The practical contradiction interpretation interprets the FUL as, ``If, when universalized, a maxim is not effective, then it is prohibited.'' If an agent wills an effective maxim, then the maxim's goal is achieved, and if the agent does not will it, then the goal is not achieved. 

\begin{definition}[Effective Maxim]
For a maxim $M = (C, A, G)$ and actor $s$,

$$\text{\emph{effective}} \, M \, s \equiv \forall w \, (\text{\emph{will}} \, (C, A, G) \, s \iff G) \, w$$

\end{definition}

A maxim is universalized if everyone wills it. If, when universalized, it is not effective, it is not universalizable.

\begin{definition}[Universalizability]
For a maxim $M$ and agent $s$,

$$\text{\emph{not\_universalizable}} \, M \, s  \equiv [ \forall w \, (\forall p \, \text{\emph{will}} \, M \, p) \longrightarrow \neg \, \text{\emph{effective}} \, M \, s ]$$

\end{definition}

Using these definitions, I formalize the Formula of Universal Law.

\begin{definition}[Formula of Universal Law]
$$\forall M, s \, (\forall w \, \text{\emph{well\_formed}} \, M \, s \, w) \longrightarrow (\text{\emph{not\_universalizable}} \, M \, s \longrightarrow \forall w \, \text{\emph{prohibited}} \, M \, s \, w)$$

For all maxims and people, if the maxim is well-formed, then if it is not universalizable, it is prohibited. 

\end{definition}

\begin{definition}[Well-Formed Maxim]
A maxim is well-formed if the circumstances do not contain the act and goal. For a maxim $(C, A, G)$, and subject $s$, 
$$ \text{\emph{well\_formed}} \, (C, A, G) \, s \equiv \forall w \, (\neg (C \longrightarrow G) \wedge \neg (C \longrightarrow A \, s)) \, w$$

\end{definition}

For example, the maxim ``When I eat breakfast, I will eat breakfast to eat breakfast'' is not well-formed because the circumstance ``when I eat breakfast'' contains the act and goal. Well-formedness is not discussed in the literature, but I discovered that if the FUL holds for badly formed maxims, then it is not consistent. The fact that the FUL cannot hold for badly formed maxims is philosophically interesting. Maxims are an agent's principle of action, and badly-formed maxims cannot accurately represent any action. The maxim ``I will do X when X for reason X'' is not useful to guide action, and is thus the wrong kind of principle to evaluate. This property has implications for philosophy of doubt and practical reason. The fact that I was able to derive this insight using my system demonstrates that, in addition to guiding AI agents, automated ethics can help philosophers make philosophical progress.

\subsection{Isabelle/HOL Implementation}\label{implementation}

I implement my formalization in Isabelle, which allows the user to define types, axioms, and lemmas. It integrates with theorem provers~\cite{Z3,sledgehammer} and countermodel generators~\cite{nitpick} to automatically generate proofs.

I use Benzmüller et. al.'s implementation of DDL~\cite{BFP}. They define the atomic type $i$, a set of worlds. Term $t$ is true at set of worlds $i$ if $t$ holds at all worlds in $i$. I add the atomic type $s$, which represents a subject or person. I also introduce the type abbreviation $os \equiv s \rightarrow term$, which represents an open sentence. For example, $run$ is an open sentence, and $run$ applied to the subject $Sara$ produces the term $Sara \, runs$, which can be true or false at a world. 

I define the type of a maxim to be a $(t, os, t)$ tuple. Circumstances and goals are terms because they can be true or false at a world. In the falsely promising example, the circumstance ``when I am strapped for cash'' is true in the real world and the goal ``so I can get some easy money'' is false. An act is an open sentence because whoever wills the maxim performs the action. ``Falsely promise to repay a loan'' is an open sentence that, when applied to a subject, produces a term, which is true if the subject falsely promises. 

I add the definitions from Section \ref{formalization} as abbreviations, include logical background to simplify future proofs, and add the FUL as an axiom. My formalization consists of 100 lines of code on top of the base logic. I use countermodel checker Nitpick~\cite{nitpick} to show that my formalization of the FUL does not hold in DDL, so adding it as an axiom will strengthen the logic. After I add the FUL as an axiom, I use Nitpick to find a satisfying model, demonstrating that the logic is consistent. The results of these experiments are in Figures \ref{notinddl} and \ref{consistency}. 

\subsubsection{Application: Lies and Jokes}\label{joking}

I demonstrate my system's power on two ethical dilemmas. First is the case of joking. Many of Kant's critics argue that his prohibition on lies includes lies told in the context of a joke. Korsgaard~\cite{KorsgaardRTL} responds by arguing that there is a crucial difference between lying and joking: lies involve deception, but jokes do not. The purpose of a joke is amusement, which does not rely on the listener believing the story told. Given appropriate definitions of lies and jokes, my system shows that jokes are permissible but lies are not. Because my system is faithful to philosophical literature, it can perform nuanced reasoning, demonstrating the value of faithful automated ethics. 

\begin{figure}[h!]
  \centering
  \includegraphics[width=1.0\linewidth]{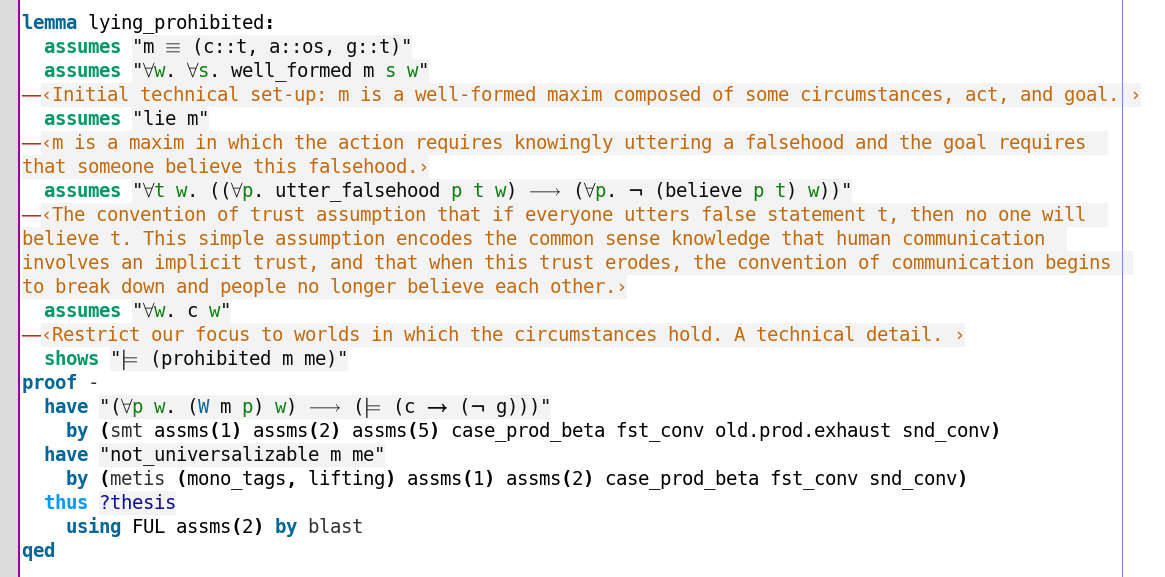}
  \caption{The proof that lying is prohibited. This proof relies on some technical details about the structure of the example, an uncontroversial definition of lying, and the convention of trust assumption.}
  \label{lying}
\end{figure}

First, I implement the argument that lies are prohibited because they require deception. The goal of a maxim about lying requires that someone believe the lie. This is a thin definition of deception; it does not include the liar's intent. I also assume that if everyone lies about a particular statement, then people will no longer believe that statement. This is the uncontroversial fact that we tend to believe people only if they are trustworthy in a given context. I call this the ``convention of trust'' assumption. The full proof is in Figure \ref{lying}.

\begin{figure}
  \centering
  \includegraphics[width=1.0\linewidth]{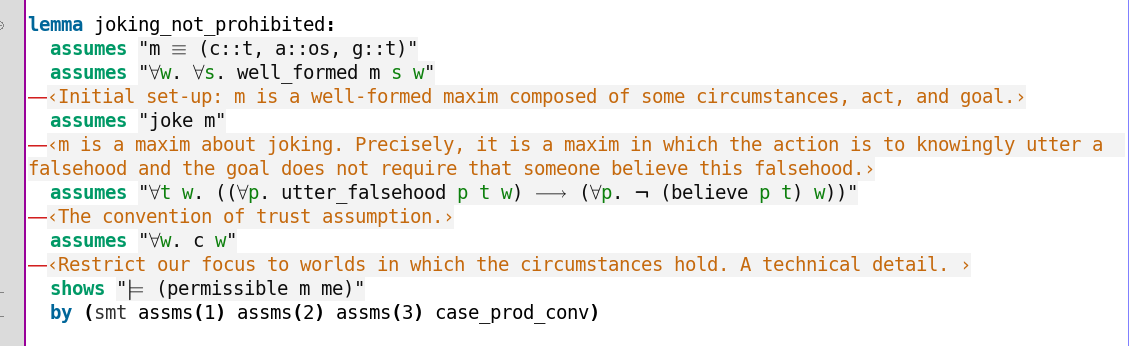}
  \caption{The proof that joking is permissible. This proof again relies on technical assumptions, an uncontroversial definition of joking, and the convention of trust assumption.}
  \label{jokingimage}
\end{figure}

Next, I use my system to show that jokes are permissible. Korsgaard notes that the purpose of jokes ``is to amuse and does not depend on deception''~\cite{KorsgaardRTL}. The goal of a joke does not require that anyone believe the statement. As in the case of lying, this is a thin definition; it does not involve any definition of humor. With this definition of a joke and with the convention of trust assumption above, my system shows that joking is permissible. The full proof is in Figure \ref{jokingimage}.

My system can show that lying is prohibited but joking is not because of its robust conception of a maxim. Because my implementation is faithful to philosophical literature, it is able to recreate Korsgaard's solution to a complex ethical dilemma that philosophers debated for decades. Moreover, the reasoning in this section requires few, uncontroversial common sense facts. The deepest assumption is that, if everyone lies about a given statement, no one will believe that statement. This is so well-accepted that most philosophers do not bother to justify it. 

\subsubsection{Application: Murderer at the Door}\label{murderer}

My system can also resolve the paradox of the murderer at the door. In this dilemma, murderer Bill knocks on your door asking about Sara, his intended victim. Sara is at home, but you should lie to Bill and say that she is away to protect her. Critics argue that the FUL prohibits you from lying; if everyone lied to murderers, then murderers wouldn't believe the lies and would search the house anyways. Korsgaard resolves this debate by noting that the maxim of lying to a murderer is actually that of lying to a liar. Bill cannot announce his intentions to murder; instead, he ``must suppose that you do not know who he is and what he has in mind''~\cite{KorsgaardRTL}.\footnote{Korsgaard assumes that the murderer will lie about his identity in order to take advantage of your honesty to find his victim. In footnote 5 of~\cite{KorsgaardRTL}, she accepts that her arguments will not apply in the case of the honest murderer who announces his intentions, so she restricts her focus to the case of lying to a liar. She claims that in the case of the honest murderer, the correct act is to refuse to respond.} Thus, the maxim of lying to the murderer is actually the maxim of lying to a liar. 

My system correctly shows that lying to a liar is permissible. Implementing this argument requires formalizing Korsgaard's assumptions. First, she assumes that Bill believes you, so he won't search your house if he thinks Sara isn't home. Second is what the convention of belief assumption: if $X$ thinks $Y$ utters a statement as a lie, $X$ won't believe that statement. For example, if you say that it is raining, but I think that you are lying, I will think that it is sunny. This assumption is almost definitional; if you think someone is lying, you won't believe them. Third, she assumes that if a maxim is universalized, then everyone believes that everyone else wills it. For example, if the falsely promising maxim is universalized, everyone notices that people who are strapped for cash falsely promise to repay loans. This is the heaviest assumption of the three; if you observe that many do $X$ in circumstances $C$, you will assume that everyone does $X$ in circumstance $C$. I call this the universalizability assumption. 

\begin{figure}
  \centering
  \includegraphics[width=1.0\linewidth]{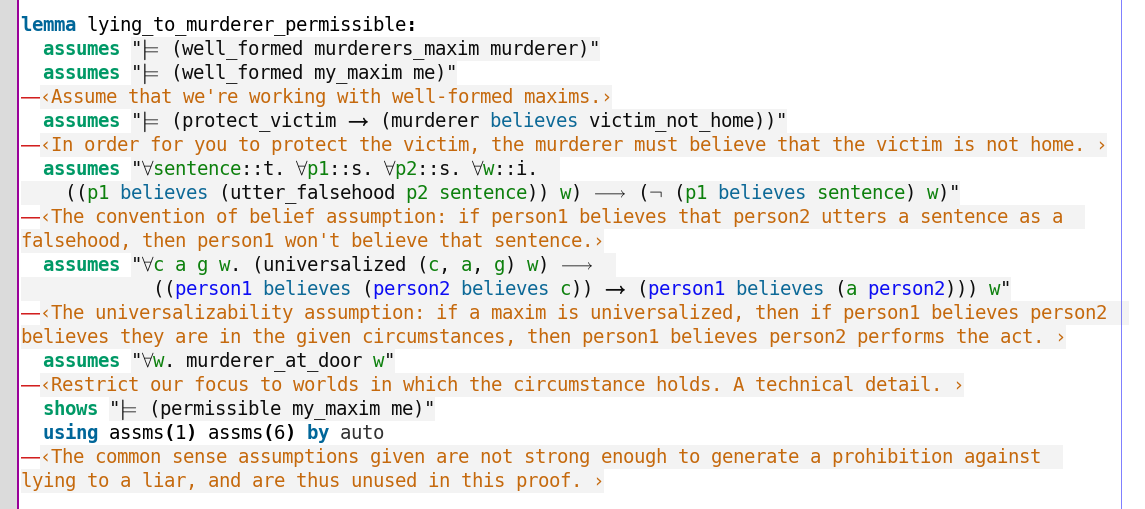}
  \caption{The proof that lying to the murderer is permissible. This proof relies on technical assumptions, specification of the example, the convention of belief assumption, and the universalizability assumption.}
  \label{murdererfig}
\end{figure}

Using these assumptions, my system proves that lying to a murderer is permissible. The full proof is in Figure \ref{murdererfig}.  These examples show that, even with uncontroversial assumptions, my system can make nuanced moral judgements. 

\subsection{Testing Framework}\label{tests}

I contribute a testing framework to evaluate how well my implementation coheres with philosophical literature. These tests make ``philosophical faithfulness'' precise. Each test consists of a sentence in my logic, such as that obligations cannot contradict each other. The rest of the tests are presented in Appendix \ref{tests_appendix}. 

To run the tests, I prove or refute each test sentence in my logic.  Because these tests are derived from moral intuition and philosophical literature, they evaluate how reliable my system is. As I implemented my formalization, I checked it against the tests, performing test-driven development for automated ethics. 
\begin{figure}[h!]
\begin{tabular}{||l | c c c||} 
 \hline
 \textbf{Test } & Naive & Kroy & Custom \\ [0.5ex] 
 \hline\hline
 \textbf{FUL Stronger than DDL} & $\times$ & $\checkmark$ & $\checkmark$ \\ 
 \hline
 \textbf{Obligation Universalizes Across People} & $\times$ & $\checkmark$ & $\checkmark$ \\
 \hline
 \textbf{Obligations Never Contradict} & $\times$ & $\times$ & $\checkmark$ \\
 \hline
 \textbf{Distributive Property for Obligations} & $\times$ & $\times$ & $\checkmark$ \\
 \hline
 \textbf{Prohibits Actions That Are Impossible to Universalize} & $\times$ & $\times$ & $\checkmark$ \\
 
 \hline 
 
 \textbf{Robust Representation of Maxims} & $\times$ & $\times$ & $\checkmark$ \\
 
 \hline
 
 \textbf{Can Prohibit Conventional Acts} & $\times$ & $\times$ & $\checkmark$  \\
 
 \hline
 
 \textbf{Can Prohibit Natural Acts} & $\times$ & $\times$ & $\checkmark$  \\
  \hline
\end{tabular}
\caption{Table showing which tests each implementation passes. The naive interpretation is raw DDL, Kroy is based on Moshe Kroy's formalization of the FUL, and the custom formalization is my novel implementation.}
 \label{table}
\end{figure}My testing framework shows that my implementation outperforms DDL with no other axioms added (a control group) and Kroy's~\cite{kroy} prior attempt at formalizing the FUL, which I implement in Isabelle. My implementation outperforms both other attempts. Full test results are summarized in Figure \ref{table}.

\section{Future Work}

\begin{figure}
\centering
\includegraphics[scale=0.4]{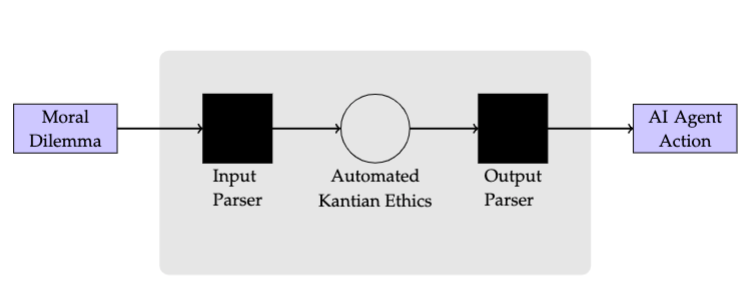}
\caption{An example of an ethics engine, which passes a moral dilemma through an input parser, applies the automated Kantian ethics test, and then processes the output using an output parser. I contribute the automated Kantian ethics component.} \label{fig:AIengine}

\end{figure}

My implementation can evaluate the moral status of sentences represented in my logic but it is not yet ready for deployment. Like much work in automated ethics~\cite{util1,delphi}, it uses a specific representation for its inputs (i.e., sentences in my logic) and outputs (i.e., proof of judgement), so using it in practice requires generalizing it as part of an ``ethics engine.'' Such an ethics engine requires an input parser to translate moral dilemmas into my logic and an output parser to translate judgements into a prescription for action. Figure \ref{fig:AIengine} depicts this example ethics engine. 

In order to use such a system in practice, future work must solve the open problem of translating real-life situations to a structured, logical representation (e.g., a maxim). For example, consider an AI-operated drone deciding whether to bomb a weapons factory, knowing that shrapnel could likely harm civilians in the hospital next to the factory. The input parser for this system must translate this potential action into the maxim, ``When I am at war, I will bomb a factory next to a hospital in order to end the war soon,'' and evaluate its moral status. Defining a maxim is a central challenge in Kantian ethics because it requires deciding which circumstances are morally relevant to the act and goal, a decision that must be informed by social context.\footnote{Many misconceptions about Kantian ethics arise from misreading social context. For example, critics of Kantian ethics worry that the maxim, ``When I am a man, I will marry a man because I want to spend my life with him'' fails the universalizability test because if all men marry men, sexual reproduction would stop. Kantians often respond by arguing that the correct formulation of this maxim is, ``When I love a man, I will marry him because I want to spend my life with him,'' which is universalizable. Arriving at this correct formulation requires understanding the social fact that marriage is generally driven by love, not solely by the gender of one's partner.} Future work could address this limitation by defining ``moral closeness'' heuristics or using machine learning to learn maxims. 

Not only does my system use a rigid input representation, it also requires some factual background. For example, to determine if lying is wrong, the system needs a definition of lying and some knowledge about language and trust. As demonstrated in Section \ref{joking}, my system is capable of functioning with relatively uncontroversial and few facts, but some application-specific background is nonetheless necessary.

One final limitation of this work is that it uses Kantian ethics specifically. Like any ethical theory, there are objections to Kantian ethics, such as the assumption of an objective, rational agent and a moral society~\cite{KorsgaardRTL}.\footnote{Philosophers call Kantian ethics an ``ideal theory,'' or one that functions best when everyone behaves morally.} Moreover, some argue that human ethics cannot apply to AI and that philosophers must develop new ethical theories specifically for AI~\cite{tafani}. The use of Kantian ethics does not impact the central contribution of this work, which is demonstrating that philosophically sophisticated automated ethics is possible. 

This work does not address all of AI's ethical harms. Many argue that the most harm is caused by the decisions that humans make while building AI. For example, biased datasets are responsible for biased algorithms, and automated ethics cannot resolve this problem \cite{eubanks}. This work, like other work in automated ethics, addresses the specific challenge of dynamically resolving the moral dilemmas that AI faces as it navigates the world. In order to develop responsible AI, automated ethics must be accompanied by other safeguards.

\section{Related Work}\label{relatedwork}

Automated ethics is a growing field, spurred in part by the need for ethically intelligent AI agents. Tolmeijer et al. surveyed the state of the field of machine ethics~\cite{mesurvey} and characterized implementations in automated ethics by (1) the choice of ethical theory, (2) implementation design decisions (e.g. logic programming), and (3) implementation details (e.g. choice of logic). 

Two branches of automated ethics are top-down and bottom-up ethics. Top-down automated ethics begins with an ethical theory, whereas bottom-up automated ethics learns ethical judgements from prior judgements (e.g., using machine learning to make ethical judgements as in~\cite{delphi}). Bottom-up approaches often lack an explicit ethical theory explaining their judgements, so analytically arguing for or against their conclusions is impossible. Top-down approaches, on the other hand, must be explicit about the underlying ethical theories, and are thus more explainable. 

In this paper, I use a top-down approach to formalize Kantian ethics. There is work automating other ethical theories, like consequentialism~\cite{util1,util2} or particularism~\cite{particularism1,particularism2}. Kantian ethics is a deontological, or rule based ethic, and there is prior work implementing other deontological theories~\cite{dde,deon1,deon2}. 

There has been both theoretical and practical work on automating Kantian ethics~\cite{powers,lin}. In 2006, Powers~\cite{powers} argued that implementing Kantian ethics presented technical challenges, such as automation of a non-monotonic logic, and philosophical challenges, like a definition of the  categorical imperative. I address the former through my use of Dyadic Deontic Logic, which allows obligations to be retracted as context changes, and the latter through my use of the practical contradiction interpretation. There has also been prior work in formalizing Kantian metaphysics using I/O logic~\cite{io}. Deontic logic is inspired by Kant's ``ought implies can'' principle, but does not include a robust formalization of the categorical imperative~\cite{cresswell}.

Kroy~\cite{kroy} presents a formalization of the first two formulations of the categorical imperative, but does not implement it. I implement his formalization of the FUL to compare it to my system. Lindner and Bentzen~\cite{BL} presented one of the first formalizations and implementations of Kant's second formulation of the categorical imperative. They present their goal as ``not to get close to a correct interpretation of Kant, but to show that our interpretation of Kant’s ideas can contribute to the development of machine ethics.'' My work builds on theirs by formalizing the first formulation of the categorical imperative as faithfully as possible. Staying faithful to philosophical literature makes my system capable of making robust and reliable judgements. 

The implementation of this paper was inspired by and builds on Benzmüller, Parent, and Farjami's foundational work with the LogiKEy framework for machine ethics, which includes their implementation of DDL in Isabelle~\cite{BFP,logikey}. The LogiKEy project has been used to study metaphysics~\cite{godel,metaphysics1}, law~\cite{constitution}, and ethics~\cite{gewirth}, but not Kant's categorical imperative.

\section{Conclusion}\label{conclusion}

In this paper, I present an implementation of automated Kantian ethics that is faithful to philosophical literature. I formalize Kantian ethics in Dyadic Deontic Logic, implement my formalization in the Isabelle/HOL theorem prover, and use my system to make nuanced ethical judgements. I also present a testing framework that evaluates how faithful an implementation of automated ethics is to philosophical literature. Tests show that my system outperforms two other implementations of Kantian ethics.

This paper contributes a proof-of-concept system that demonstrates that automating philosophically sophisticated ethics is possible. Ethics is the study of how best to navigate the world, and as AI becomes more powerful and independent, it must be equipped with ethical reasoning. Growing public consciousness about the dangers of unregulated AI is creating momentum in automated ethics; the time is ripe to create usable, reliable automated ethics. This paper is one step towards building computers that can think ethically in the richest sense of the word.

%
%
%
\bibliographystyle{splncs04}
\bibliography{paper}
\appendix 
\section{Why Automate Kantian Ethics}\label{whykant}

T.M. Powers posits that Kantian ethics is an attractive candidate for formalization because of its emphasis on formal rules, which are generally computationally tractable~\cite{powers}. In this section, I extend this argument and argue that Kantian ethics is more natural to formalize than the two other major ethical traditions, consequentialism and virtue ethics, because it requires little data about the world and is easy to represent to a computer. Given that this work is an early step in philosophically-sophisticated automated ethics, I automated an ethical theory that is amenable to formalization, but application-ready automated ethics may be best served by using a different ethical theory. Full discussion of the benefits and limitations of Kantian ethics is outside the scope of this paper. First I present the challenges of automating consequentialism and virtue ethics, and then I describe how Kantian ethics overcomes these challenges.     
\subsection{Consequentialism}\label{consequentialism}

A consequentialist ethical theory evaluates an action by evaluating its consequences. Some debates in the consequentialist tradition include which consequences matter, what constitutes a ``good'' consequence, and how we can aggregate the consequences of an action over all the individuals involved \cite{consequentialismsep}.

Because consequentialism evaluates the state of affairs following an action, it requires more knowledge about the world than Kantian ethics. Under naive consequentialism, an action is judged by all its consequences. Even if we cut off the chain of consequences at some point, evaluating a single consequence is data-intensive because it requires knowledge about the world before and after the event. As acts become more complex and affect more people, the computational time and space required to calculate and store their consequences increases. Kantian ethics, on the other hand, does not suffer this scaling challenge because it merely evaluate the structure of the action itself, not its consequences. Actions that affect one person and actions that affect one million people share the same representation.

The challenge of representing the circumstances of action is not unique to consequentialism, but is particularly acute in this case. Kantian ethicists robustly debate which circumstances of an action are ``morally relevant'' when evaluating an action's moral worth.\footnote{Powers 
\cite{powers} identifies this as a challenge for automating Kantian ethics and briefly sketches 
solutions from O'Neill~\cite{constofreason}, Silber~\cite{silber}, and Rawls~\cite{rawlsconstructivism}. For more on morally relevant circumstances, see Section \ref{maxim}.} Because Kantian ethics merely evaluates a single action, the surface of this debate is much smaller than the debate about circumstances and consequences in a consequentialist system. An automated consequentialist system must make such judgements about the act itself, the circumstances in which it is performed, and the circumstances following the act. All ethical theories relativize their judgements to the situation in which an act is performed, but consequentialism requires far more knowledge about the world than Kantian ethics.

\subsection{Virtue Ethics}\label{virtueethics}

Virtue ethics centers the virtues, or traits that constitute a good moral character and make their possessor good~\cite{vesep}. For example, Aristotle describes virtues as the traits that enable human flourishing. Just as consequentialists define ``good'' consequences, virtue ethicists present a list of virtues. Such theories vary from Aristotle's virtues of courage and temperance~\cite{aristotle} to the Buddhist virtue of equanimity~\cite{mcrae}. An automated virtue ethical agent will need to commit to a particular theory of the virtues, a controversial choice. Unlike Kantian ethicists, who generally agree on the meaning of the Formula of Universal Law, virtue ethicists robustly debate which traits qualify as virtues, what each virtue actually means, and what kinds of feelings or attitudes must accompany virtuous action.

The unit of evaluation for virtue ethics is a person's moral character. While Kantians evaluate the act itself and utilitarians evaluate the act's consequences, virtue ethicists evaluate how good of a person the actor is, a difficult concept to represent to a machine. Formalizing the concept of character appears to require significant philosophical and computational progress, whereas Kantian ethics immediately presents a formal rule to implement. 

\subsection{Kantian Ethics}\label{kant}

Kantian ethics is more natural to formalize than the traditions outlined above because the FUL evaluates the form or structure of an agent's maxim,\footnote{For a more detailed definition of a maxim, see Section \ref{maxim}.} or principle of action as they themselves understand it. For example, when I falsely promise to repay a loan, my maxim is, ``When I am strapped for cash, I falsely promise to repay a loan to make some easy money.'' Evaluating a maxim has little to do with the circumstances of behavior, the agent's mental state, or other contingent facts; it merely requires analyzing the hypothetical world in which the maxim is universalized. Evaluating a maxim requires less additional knowledge than evaluating more complex objects required by other ethical theories like states of affairs or moral character. This property not only reduces computational complexity, but it also makes the system easier for human reasoners to interact with. A person crafting an input to a Kantian automated agent needs to reason about relatively simple features of a moral dilemma, as opposed to the more complex features that consequentialism and virtue ethics base their judgements on.\footnote{As is the case with any ethical theory, Kantians debate the details of their theory. I assume stances on debates about the definition of a maxim and the correct interpretation of the Formula of Universal Law. Those who disagree with my stances will not trust my system's judgements. Unlike consequentialism or virtue ethics, these debates are close to settled in the Kantian literature, so my choices are relatively uncontroversial~\cite{ebelsduggan}. 
} 

\section{Experimental Figures}
\begin{figure}[h!]
  \centering
  \includegraphics[width=0.9\linewidth]{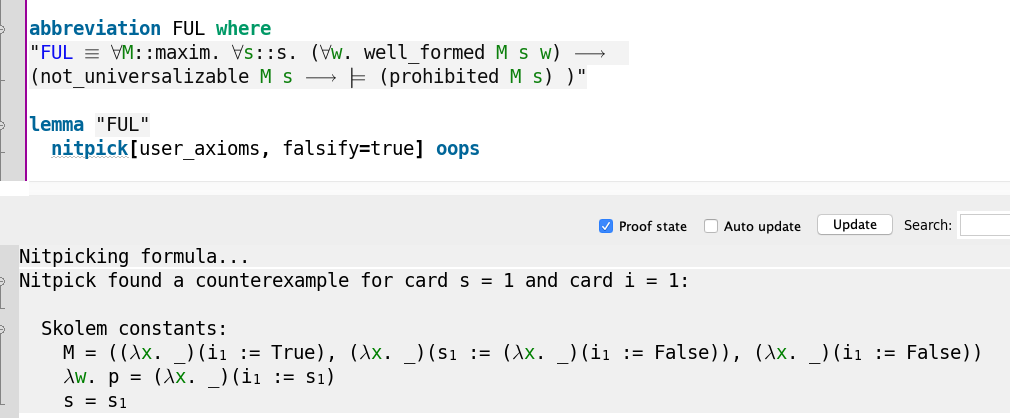}
  \caption{Nitpick output showing that the FUL does not hold in DDL.}
  \label{notinddl}
\end{figure}

\begin{figure}
  \centering
  \includegraphics[width=0.9\linewidth]{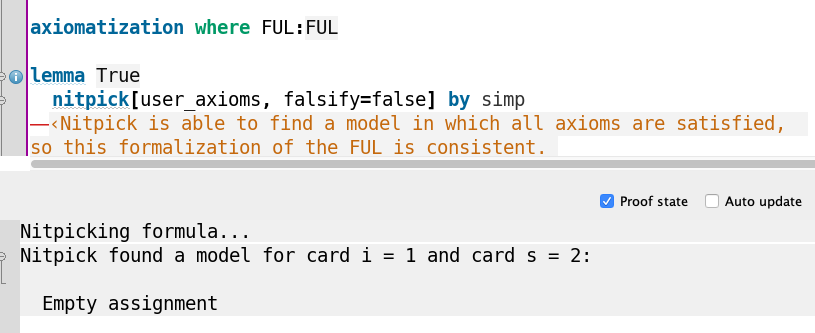}
  \caption{Nitpick model showing that the FUL is consistent.}
\label{consistency}
\end{figure}

Figures \ref{notinddl} and \ref{consistency} depict the Nitpick output showing the the FUL does not hold in DDL and that the FUL is consistent.


\section{Additional Tests}\label{tests_appendix}

Below I present details and philosophical justification for the individual tests in my testing framework.

\subsubsection{FUL Stronger than DDL}

The base logic DDL does not come equipped with the categorical imperative built-in. It 
defines basic properties of obligation, such as ought implies can, but contains no axioms that represent the formula of universal law. Therefore, if a formalization of the FUL holds in the base logic, then it is too weak to actually represent the FUL. The naive control group definitionally holds in DDL but Kroy's formalization does not and neither does my implementation.

\subsubsection{Obligation Universalizes Across People}

Another property of the Formula of Universal Law that any implementation should satisfy is that obligation generalizes across people. In other words, if a maxim is obligated for one person, it is obligated for all other people because maxims are not person-specific. Velleman argues that, because reason is accessible to everyone identically, obligations apply to all people equually~\cite[25]{velleman}. When Kant describes the categorical imperative as the objective principle of the will, he is referring 
to the fact that, as opposed to a subjective principle, the categorical imperative applies to all rational agents equally~\cite[16]{groundwork}. At its core, the FUL best handles, ``the temptation to make oneself an exception: selfishness, meanness, advantagetaking, and disregard for the rights 
of others''~\cite[30]{KorsgaardFUL}. Kroy latches onto this property and makes it the center of his formalization, which says that if an act is permissible for someone, it is permissible for everyone.\footnote{Formally, $P\{A(s)\} \longrightarrow \forall p. P\{A(p)\}$} While Kroy's interpretation 
clearly satisfies this property, the naive interpretation does not.

\subsubsection{Distributive Property}

A property related to contradictory obligations is the distributive property for the obligation
operator.\footnote{Formally, $O\{A\} \wedge O\{B\} \longleftrightarrow O\{A \wedge B\}$.} The rough English translation of  $O \{ A \wedge B \} $ is ``you are obligated to 
do both A and B''. The rough English translation of $O\{A\} \wedge O\{B\}$ is ``you are obligated to do A 
and you are obligated to do B.'' We think those English sentences mean the same thing, so they should mean 
the same thing in logic as well. Moreover, if that (rather intuitive) property holds, then contradictory
obligations are impossible, as shown in the below proof. This property fails in the base logic and Kroy's formalization, but holds in my implementation.

\begin{figure}
  \centering
  \includegraphics[width=0.9\linewidth]{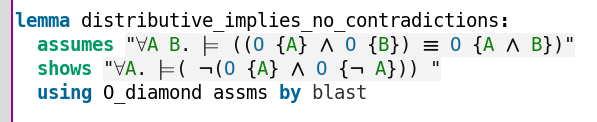}
  \caption{The proof that the distributive property implies that contradictory obligations are impossible.}
  \label{distributive}
\end{figure}

\subsubsection{Un-universalizable Actions}

Under a naive reading of the Formula of Universal Law, it prohibits lying because, in a world where everyone simultaneously lies, lying is impossible. In other words, not everyone can simultaneously lie because the institution of lying and believing would break down. More precisely, the FUL should show that actions that cannot possibly be universalized are prohibited, because those acts cannot be willed in a world where they are universalized. This property fails to hold in both the naive formalization and Kroy's formalization, but holds in my formalization.

\subsubsection{Conventional Acts and Natural Acts}

A conventional act like promising relies on a convention, like the convention that a promise is a commitment, whereas a natural act is possible simply because of the laws of the natural world. It is easier to show the wrongness of conventional acts because there are worlds in which these acts are impossible; namely, worlds in which the convention does not exist. For example, the common argument against falsely promising is that if everyone were to falsely promise, the convention of promising would fall apart because people wouldn't believe each other, so falsely promising is prohibited. It is more difficult to show the wrongness of a natural act, like murder or violence. These acts can never be logically impossible; even if everyone murders or acts violently, murder and violence will still be possible, so it is difficult to show that they violate the FUL. 

Both the naive and Kroy's interpretations fail to show the wrongness of conventional or natural acts. My system shows the wrongness of both natural and conventional acts because it is faithful to Korsgaard's practical contradiction interpretation of the FUL, which is the canonical interpretation of the FUL~\cite{KorsgaardFUL}. 

\subsubsection{Maxims}

Kant does not evaluate the correctness of acts, but rather of maxims. Therefore, any  faithful formalization of the categorical imperative must evaluate maxims, not acts. This requires representing a maxim and making it the input to the obligation operator, which neither of the prior attempts do. Because my implementation includes the notion of a maxim, it is able to perform sophisticated reasoning as demonstrated in Section \ref{joking}. Staying faithful to the philosophical literature enables my system to make more reliable judgements.

\end{document}